\pdfoutput=1
\documentclass[letterpaper, 10 pt, conference]{ieeeconf}  

\IEEEoverridecommandlockouts                              

\usepackage{graphics}
\usepackage{bbm}
\usepackage{dsfont}
\usepackage[pdftex]{graphicx}
\usepackage{wrapfig}
\DeclareGraphicsExtensions{.pdf,.png,.jpg}
\usepackage{epsfig}
\usepackage[font={small}]{caption}
\usepackage{subcaption}
\usepackage[rightcaption]{sidecap}
\usepackage{pbox}

\usepackage{bigstrut}
\usepackage{mathtools}
\usepackage{amsmath, amssymb, amscd}
\usepackage{ wasysym } 
\usepackage{mathptmx} 
\usepackage{gensymb} 
\usepackage{nicefrac}       
\numberwithin{equation}{section} 
\usepackage{dsfont}

\DeclareMathAlphabet{\mathcal}{OMS}{lmsy}{m}{n}
\usepackage{textcomp} 

\usepackage{algorithm} 
\usepackage[noend]{algorithmic} 

\usepackage{array} 
\usepackage{tabularx}
\usepackage{multirow}
\usepackage{booktabs}
\usepackage{tabulary}

\usepackage[T1]{fontenc} 
\usepackage[utf8]{inputenc}
\usepackage[english]{babel} 
\usepackage{units}
\usepackage{bm}
\usepackage{times} 
\usepackage{xspace}
\usepackage{flushend}
\usepackage{balance} 
\usepackage{csquotes}
\usepackage{makeidx}
\usepackage{blindtext}

\let\labelindent\relax
\usepackage{enumitem}

\usepackage{ragged2e}
\usepackage{soul} 
\usepackage{subfiles} 

\usepackage[protrusion=true,expansion=true]{microtype}
\usepackage[yyyymmdd]{datetime}

\date{\protect\formatdate{1}{1}{2001}}

\usepackage{url}
\makeatletter
\g@addto@macro{\UrlBreaks}{\UrlOrds}
\makeatother
\usepackage{color}
\usepackage[usenames,dvipsnames,table,xcdraw]{xcolor}
\usepackage{pgfplots} 
\pgfplotsset{compat=newest}

\usepackage{marginnote}
\usepackage{soul} 

\usepackage{multicol}

\newcommand{\tocite}[1]{%
\textcolor{red}{[cite:\ifthenelse{\equal{#1}{}}{}{#1}?]}}

\newcommand{\ignore}[1]{}

\renewcommand{\baselinestretch}{1.0}



\usetikzlibrary{backgrounds}
\numberwithin{equation}{section} 

\tikzstyle{every node}=[font=\small]

\newcolumntype{L}[1]{>{\RaggedRight\hspace{0pt}}p{#1}}
\newcolumntype{R}[1]{>{\RaggedLeft\hspace{0pt}}p{#1}}

\newcommand{\furlp}[1]{\colorbox{blue!10}{\href{run:/home/fulong/academia/library/papers/#1.pdf}{D}}}
\newcommand{\furlb}[1]{\colorbox{blue!10}{\href{run:/home/fulong/academia/library/books/#1.pdf}{D}}}

\title{\Large \bf
Intermittent Visual Servoing: Efficiently Learning Policies\\Robust to Instrument Changes for High-precision Surgical Manipulation
}

\author{Samuel Paradis$^1$, Minho Hwang$^1$, Brijen Thananjeyan$^1$, Jeffrey Ichnowski$^1$, Daniel Seita$^1$, \\ Danyal Fer$^2$,  Thomas Low$^3$,  Joseph E. Gonzalez$^1$, Ken Goldberg$^1$
\thanks{$^{1}$University of California, Berkeley, USA.}%
\thanks{$^{2}$UC San Francisco East Bay, USA.}%
\thanks{$^{3}$SRI International, USA.}%
\thanks{Correspondence: Samuel Paradis, {\tt\small samparadis@berkeley.edu}}%
}
\begin{document}

\maketitle
\thispagestyle{empty}
\pagestyle{empty}

\begin{abstract}
Automation of surgical tasks using cable-driven robots is challenging due to backlash, hysteresis, and cable tension, and these issues are exacerbated as surgical instruments must often be changed during an operation. In this work, we propose a framework for automation of high-precision surgical tasks by learning sample efficient, accurate, closed-loop policies that operate directly on visual feedback instead of robot encoder estimates. This framework, which we call intermittent visual servoing (IVS), intermittently switches to a learned visual servo policy for high-precision segments of repetitive surgical tasks while relying on a coarse open-loop policy for the segments where precision is not necessary. To compensate for cable-related effects, we apply imitation learning to rapidly train a policy that maps images of the workspace and instrument from a top-down RGB camera to small corrective motions. We train the policy using only 180 human demonstrations that are roughly 2 seconds each. Results on a da Vinci Research Kit suggest that combining the coarse policy with half a second of corrections from the learned policy during each high-precision segment improves the success rate on the Fundamentals of Laparoscopic Surgery peg transfer task from 72.9\% to 99.2\%, 31.3\% to 99.2\%, and 47.2\% to 100.0\% for 3 instruments with differing cable-related effects. In the contexts we studied, IVS attains the highest published success rates for automated surgical peg transfer and is significantly more reliable than previous techniques when instruments are changed. Supplementary
material is available at \url{https://tinyurl.com/ivs-icra}.
\end{abstract}
\section{Introduction}

Laparoscopic surgical robots such as the da Vinci Research Kit (dVRK)~\cite{dvrk2014} are challenging to accurately control using open-loop techniques,
because of the hysteresis, cable-stretch, and complex dynamics of their cable-driven joints%
~\cite{kalman_filter_2016,miyasaka2015,hwang2020efficiently}. Furthermore, encoders are typically located at the motors, far from the joints they control, making accurate state estimation challenging. Prior work addresses these issues by learning a model of the robot's dynamics from data~\cite{hwang2020efficiently,peng2020real,thananjeyan2019safety} for accurate open-loop control or by learning control policies that directly command the robot to perform tasks~\cite{recovery-rl}. However, these approaches tend to require many training samples, which can take a long time to collect on a physical robot. Additionally, learning a model of the robot's dynamics requires accurate state estimation, which may necessitate sophisticated motion capturing techniques using fiducials~\cite{hwang2020efficiently,peng2020real}. Also, the learned dynamics models can overfit to the specific cabling properties of individual instruments (see Section~\ref{subsec:transferability}).      
Because instrument changes are commonplace within and across surgeries, control strategies must be robust to these shifts in cabling properties. 

\begin{figure}[t!]
\centering
\includegraphics[width=1.0\linewidth]{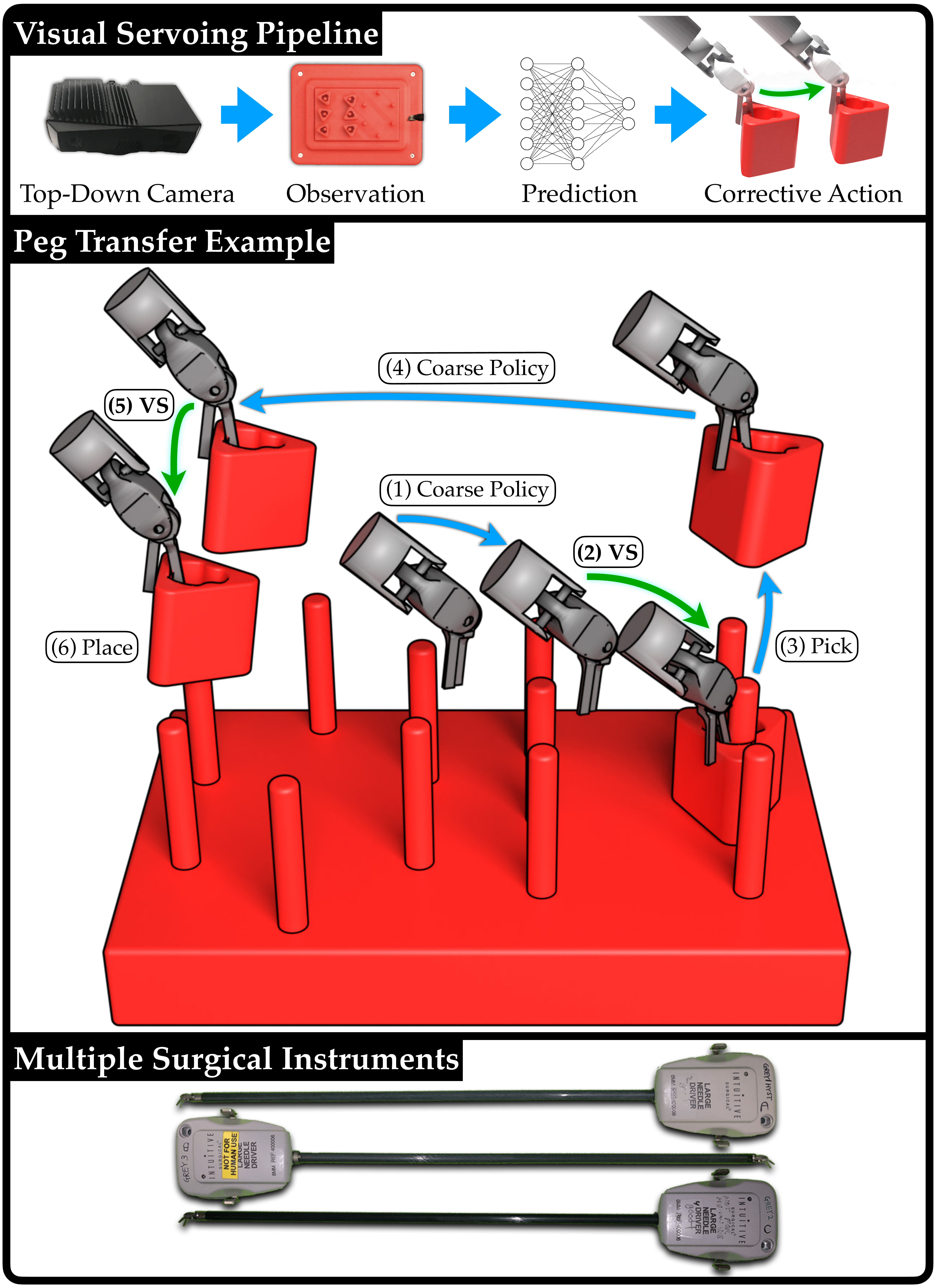}
\caption{
\textbf{Intermittent Visual Servoing (IVS).}
\textbf{Visual Servoing Pipeline:} To compensate for cable-related effects, a policy maps images of the workspace from a top-down RGB camera to corrective motions. \textbf{Peg Transfer Example:} The robot switches to a visual feedback policy for high-precision segments of the task (green), and uses a coarse policy to navigate between these segments (blue). \textbf{Multiple Surgical Instruments:} We experiment using 3 different large needle drivers, each with unique backlash, hysteresis, and cable tension.
} 
\label{fig:main}
\end{figure}

We propose a framework called intermittent visual servoing (IVS), which combines coarse planning over a robot model with learning-based, visual feedback control at segments of the task that require high precision. We use RGBD sensing to construct open-loop trajectories to track with a coarse policy, but during intermittent servoing, we only use RGB sensing, as it can capture images at a much higher frequency. Further, depth sensing requires a static environment, so using RGB allows for continuous visual servoing while the robot is still in motion. The higher capture rate of RGB imaging combined with not requiring the robot to fully stop to sense allows for 10.0 corrective updates per second during peg transfer experiments, compared with only 1.6 for RGBD servoing.

We use imitation learning (IL) to train a precise, visual feedback policy from expert demonstrations. Imitation learning is a popular approach for training control policies from demonstrations provided by a human or an algorithmic supervisor, but may require significant amounts of data~\cite{argall2009survey,imitation_survey_2018}, which is expensive in the case of a human supervisor~\cite{dagger}. To mitigate this requirement, the learning-based visual feedback policy is only trained on segments where accuracy is necessary, while we rely on a coarse open-loop policy to navigate between these segments. As a result, training the policy requires fewer demonstrations than a policy trained to perform the entire task. Because this policy is trained to directly output controls from images, it does not require explicit state estimation techniques as in prior work. While dVRK surgical instruments can have errors up to 6~mm in positioning~\cite{hwang2020efficiently}, this is sufficient for low-precision segments such as transferring a block between pegs.

\begin{figure}[t]
\centering
\includegraphics[width=1.0\linewidth]{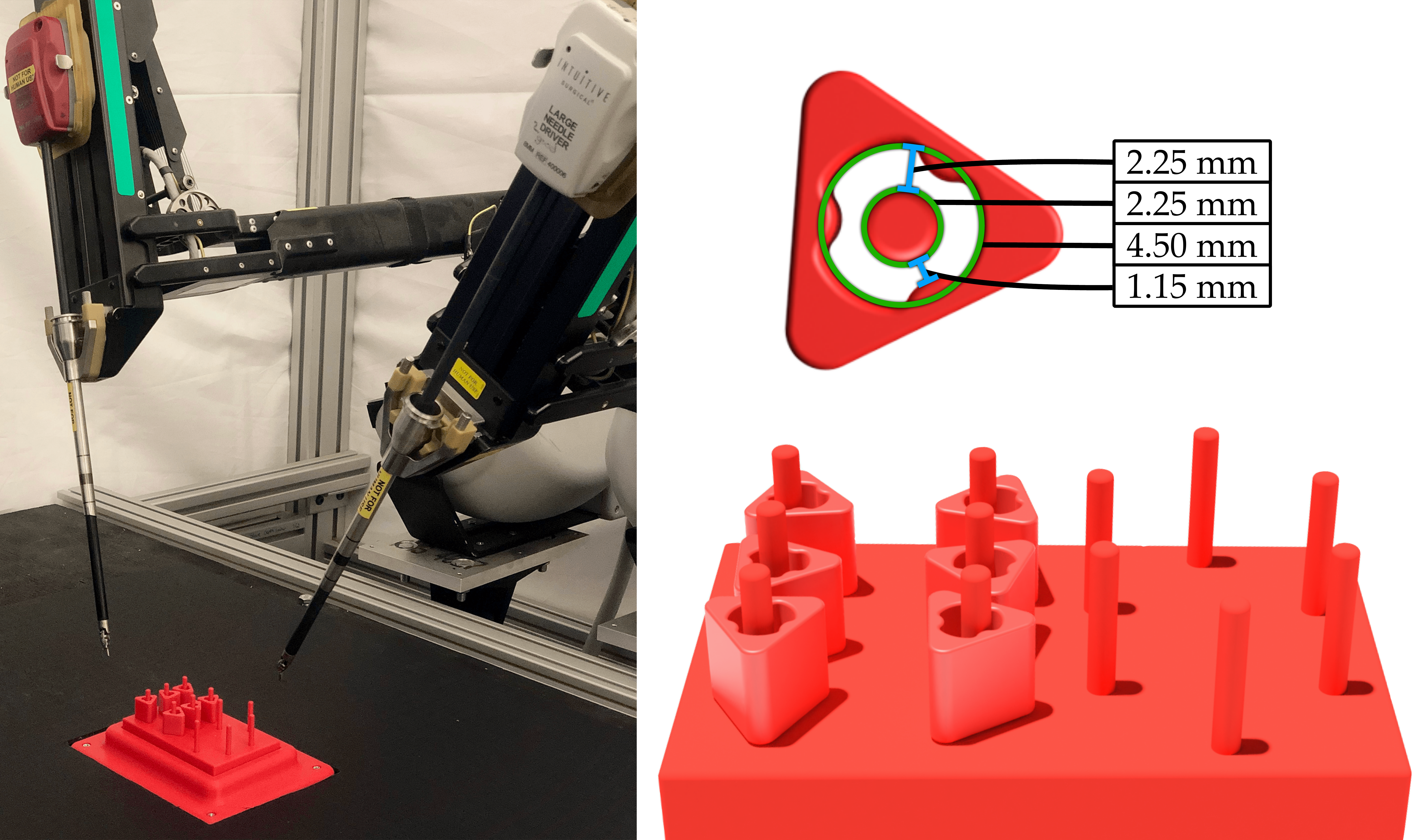}
\caption{\textbf{Robot Setup for FLS Peg Transfer.} We use Intuitive Surgical’s da Vinci Research Kit robot. We use red 3D printed blocks and a red 3D printed peg board. We use a uniformly red setup to simulate the surgical environment, where color alone may not provide sufficient signal, as surgeons rely on minute differences in color, depth and texture to complete high-precision tasks. We use a top-down camera to generate open loop trajectories (RGBD) and capture the input for visual servoing (RGB). The FLS Peg Transfer task involves transferring 6 blocks from the 6 left pegs to the 6 right pegs, and transferring them back from the right pegs to the left pegs. As each block's opening has a 4.5~mm radius, and each peg's cylindrical is 2.25~mm wide, the task requires high precision, making it a popular benchmark task for evaluating human surgeons.
} 
\label{fig:setup}
\end{figure}

We focus on the Fundamentals of Laparoscopic Surgery (FLS) peg-transfer surgeon training task, in which the surgeon must transfer 6 blocks from a set of pegs to another set of pegs, and then transfer them back (Fig.~\ref{fig:setup}). As each block's opening has a 4.5~mm radius, and each peg's cylindrical is 2.25~mm wide, the task requires high precision, making it a popular benchmark task for evaluating human surgeons~\cite{haptic-feedback-surgery-2019,rating_peg_transfer_2017,ur5-peg-transfer-2019,laparoscopic_transfer_2014,madapana2019desk, rahmantransferring}. Prior work in automating peg transfer suggests that servoing based on encoder readings cannot reliably perform this task, as positioning errors lead to failure~\cite{hwang2020efficiently, hwang2020applying}. As a result, sophisticated calibration techniques are used to correct for cabling effects of the surgical instrument during execution~\cite{hwang2020efficiently,hwang2020applying}. 
However, accuracy is only required directly before grasping or releasing the block, not during larger motions such moving with the block from one peg to another.

Switching surgical instruments during surgery is both necessary and common~\cite{transfer_1}. Depending on the type of procedure, up to four instruments may be exchanged on a single arm in rapid succession to perform a task, and this may occur multiple times over a given procedure~\cite{transfer_1}. These exchanges have been demonstrated to contribute to 10 to 30\% of total operative time, increasing patient exposure to anesthesia~\cite{transfer_2}. Additionally, each instrument is only permitted to be used for 10 operations regardless of the operation length due to potential instrument degradation and even within this permitted-use window instruments frequently fail~\cite{transfer_3,transfer_4}. Moreover, between patients, instruments must undergo high pressure, high heat sterilization that further degrades the instrument~\cite{transfer_3}. Instrument collisions during a procedure are common and can alter the cabling properties of the instrument, necessitating re-calibration in the case of automated surgery. Sophisticated, instrument-specific calibration techniques require many long trajectories of data~\cite{hwang2020efficiently}, which further increases the wear on the instrument, reduces its lifespan, and can require time during or before a surgical procedure to collect data. Therefore, developing policies that are efficiently transferable across instruments is critical to automation of surgical tasks, since instruments are exchanged frequently and instrument properties change over time with increased usage.

This paper makes the following contributions: (1) IVS, a novel deep learning framework for automation of high-precision surgical tasks, (2) experiments on the FLS peg transfer task suggesting that IVS can match state-of-the-art calibration methods in terms of accuracy while requiring significantly less training data, (3) experiments suggesting that IVS is significantly more robust to instrument changes than prior methods, maintaining accuracy despite different robot instruments having different cabling-related properties. 

\section{Related Work}\label{sec:rw}

Robot assisted surgery has been widely adopted based on pure teleoperation, as exemplified by the popularity of curricula such as the Fundamentals of Laparoscopic Surgery (FLS)~\cite{fls_1998,fls_2004}. \emph{Automating} robot surgery has proven to be difficult, and so far all standard surgical procedures using robot surgery techniques use a trained human surgeon to \emph{teleoperate} the robot arms to compensate for various forces and inaccuracies in the robot system~\cite{yip2017robot}. Automating surgical robotics using cable-driven Robotic Surgical Assistants (RSAs) such as the da Vinci~\cite{dvrk2014} or the Raven II~\cite{raven2013} is known to be a difficult problem due to backlash, hysteresis, and other errors and inaccuracies in robot execution~\cite{mahler2014case,seita_icra_2018,peng2020real}. 

\subsection{Surgical Robotics Tasks}

Automation of surgical robotics tasks has a deep history in the robotics research literature. Key applications include cutting~\cite{thananjeyan2017multilateral}, debridement~\cite{Kehoe2014,murali2015learning,seita_icra_2018}, suturing~\cite{sen2016automating,rosen_icra_suturing_2017,saeidi_suturing_icra_2019}, and more broadly manipulating and extracting needles~\cite{extraction_needles_2019,needle_insertion_deformation_2019,automated_needle_pickup_2018}. 

Several research groups have explored automating the FLS peg transfer task for robot surgery, most prominently Rosen and Ji~\cite{auto_peg_transfer_2015} and Hwang~et~al.~\cite{hwang2020applying,hwang2020efficiently}. These works, while producing effective and reliable results, require sophisticated visual servoing or calibration techniques, and the resulting systems are instrument-specific. In contrast, we focus on a system that does not require accurate calibration and transfers across a variety of surgical instruments.

\subsection{Visual Servoing for High Precision Tasks}

Visual servoing has a rich history in robotics~\cite{tutorial_visuo_servo_1996,Kragic02surveyon}. Classical visual servoing mechanisms typically use domain-specific knowledge in the form of image features or system dynamics~\cite{basic_approaches_2006,caron_servoing_2013}. In recent years, data-driven approaches to visual servoing have gained in popularity as a way to generalize from patterns in larger training datasets. For example, approaches such as Levine~et~al.~\cite{levine2017} and Kalashnikov~et~al.~\cite{QT-Opt} train visual servoing policies for grasping based on months of nonstop data collection across a suite of robot arms. Other approaches for learning visual servoing include Lee~et~al.~\cite{lee_servoing_2017}, who use reinforcement learning and predictive dynamics for target following, Saxena~et~al.~\cite{cnns_end2end_visual_servoing_2017} for servoing of quadrotors, and Bateux~et~al.~\cite{deep_nn_visual_servoing_2018} for repositioning robots from target images.

In this work, to facilitate rapid instrument changes, it is infeasible to obtain massive datasets by running the da Vinci repeatedly for each new instrument, hence we prioritize obtaining high-quality demonstrations~\cite{argall2009survey,imitation_survey_2018} at the critical moments of when the robot inserts or removes blocks from pegs. This enables the system to rapidly learn a robust policy for the region of interest, while relying on a coarse, open-loop policy otherwise. Coarse-to-fine control architectures combining geometric planners with adaptive error correction strategies have a long history in robotics~\cite{fine-motion-old, 100240, rauch2019learning,taylor1976synthesis,lozano1976design}, with works such as Lozano-Pérez et al.~\cite{fine-motion-old} studying the combination of geometric task descriptions with sensing and error correction for compliant motions. We extend the work of Lee et al.~\cite{guapo}, who use a model-based planner for moving a robot arm in free space, and reinforcement learning for learning an insertion policy when the gripper is near the region of interest.



\section{Problem Definition}\label{sec:problem-def}

We focus on the FLS peg transfer task, using the setup in Hwang et al.~\cite{hwang2020efficiently}, which uses red 3D printed blocks and a red 3D printed pegboard (see Fig.~\ref{fig:setup}). In real surgical environments, blood is common, so surgeons rely on minute differences in color, depth, and texture to complete high-precision tasks. We use a uniformly red pegboard setup, so the resulting difficulty in sensing the state in such an environment more accurately reflects the surgical setting. The task involves transferring $6$ blocks from the $6$ left pegs to the $6$ right pegs, and transferring them back from the right pegs to the left pegs (see Fig.~\ref{fig:setup}). As in Hwang~et~al.~\cite{hwang2020applying,hwang2020efficiently}, we focus on the single-arm version of the task.




We define the peg transfer task as consisting of a series of smaller subtasks, with the following success criteria:

\begin{description}
    \item[Pick:] the robot grasps a block and lifts it off the pegboard.
    \item[Place:] the robot securely places a block around a target peg.
\end{description}


We define a \textbf{transfer} as a successful pick followed by a successful place.
A \textbf{trajectory} consists of a single instance of any of the two subtasks in action. A single \textbf{trial} of the peg transfer task initially consists of 6 blocks starting on one side of the peg board, each with random configurations. A successful trial without failures consists of 6 transfers to move all 6 blocks to the other side of the board, and then 6 more transfers to move the blocks back to the original side of the peg board. A trial can have fewer than 12 transfers if failures occur during the process.

\section{IVS: Method}\label{sec:method}



\subsection{Subtask Segmentation and Policy Design}
Due to cabling effects, tracking an open-loop trajectory to pick or place targets using the robot's encoder estimates may result in positioning errors; we thus propose decomposing subtasks into 3 phases: (1) an open-loop \emph{approach} phase, (2) a closed-loop visual-servoing \emph{correction} phase, and (3) an open-loop \emph{completion} phase.
The open-loop phases are executed by a coarse policy $\pi_{\rm coarse}$ that tracks predefined trajectories using the robot's odometry. The closed-loop phases are executed by a learned, visual feedback policy $\pi_{\rm VS}$ that corrects the robot's position for the subsequent completion motion. At time $t$, the executed policy outputs an action vector $a_t$, as well as a termination signal $\phi_t \in \{0,1\}$, which signals to the system to switch to the next segment.

\subsubsection*{Pick Subtask}
The first segment uses an open-loop policy $\pi_{\rm coarse}^{\rm pick}$ to execute a trajectory to a position above the target grasp (\emph{approach}). After this motion, a visual feedback policy $\pi_{\rm VS}^{\rm pick}$ takes over to correct for positioning errors (\emph{correction}). Once corrected, the robot again executes $\pi_{\rm coarse}^{\rm pick}$ to perform a predefined grasping motion relative to its current pose (\emph{completion}).


\subsubsection*{Place Subtask}
Similar to block picking, an open-loop policy $\pi_{\rm coarse}^{\rm place}$ executes a trajectory to a position above the target placement (\emph{approach}). After this motion, a visual feedback policy $\pi_{\rm VS}^{\rm place}$ takes over to correct for positioning errors (\emph{correction}). Once corrected, the robot opens its jaws, resulting in the block dropping onto the peg (\emph{completion}).



\subsection{Fine-Policy Data Collection}
We collect demonstrations from a human teleoperator to generate a dataset to train a neural network for a fine policy. 
We collect 15 trajectories on each of the 12 pegs for both subtasks, resulting in 180 transfers, and 360 expert trajectories. Each trajectory consists of a small corrective motion, as the teleoperator navigates the end-effector from a starting position to the goal position. For picks, the goal position is directly above the optimal pick spot.  For places, the goal position is such that the center of the block aligns with the center of the target peg. The starting position of each attempt is a random position within 5~mm of the goal position.
Further, due to the size of the blocks, small segments of irrelevant blocks may be visible after data preprocessing (see Sec.~\ref{subsubsec:image-filtering}). To capture this data property, while collecting trajectories for a given target peg, we populate neighboring pegs with blocks. Then, prior to each attempt, we randomize the configurations of the neighboring blocks. 

For each demonstration, we capture a top-down RGB image $I^{\rm raw}\in\mathbb{R}^{1200\times1900\times3}$ and end-effector position $p\in\mathbb{R}^{2}$ in the robot's base frame estimated from encoder values at 5~Hz. We do not record the $z$ coordinate, because the correction phase for both subtasks will be performed in an plane with a fixed $z$ coordinate. While the recorded end-effector position has errors due to cabling properties~\cite{hwang2020efficiently,peng2020real,hwang2020applying}, we demonstrate empirically that the high-frequency visual feedback policy trained from supervision extracted from these estimates is reliable (Sec.~\ref{sec:exps}). Each demonstration is a raw trajectory:
\begin{align*}
    \mathcal{T}^{\rm raw} &= \left\{(I_t^{\rm raw}, p_t)\right\}_{t=0}^T
\end{align*}
The pick dataset $\mathcal{D}_{\rm pick} = \left\{\mathcal{T}^{\rm raw}_{\rm pick, i}\right\}_{i=1}^{N_{\rm pick}}$ consists of 2400 datapoints, and the place dataset $\mathcal{D}_{\rm place} = \left\{\mathcal{T}_{\rm place, i}^{\rm raw}\right\}_{i=1}^{N_{\rm place}}$ contains 1804 datapoints. The corrective pick trajectories are slighter longer than corrective place trajectories, resulting in 3.3 additional datapoints per demonstration.

\subsection{Preprocessing of Visual Feedback Policy Training Data}\label{subsec:data-filtering}
\subsubsection{Image Filtering}\label{subsubsec:image-filtering}
\begin{figure}[t]
\centering
\includegraphics[width=1.0\linewidth]{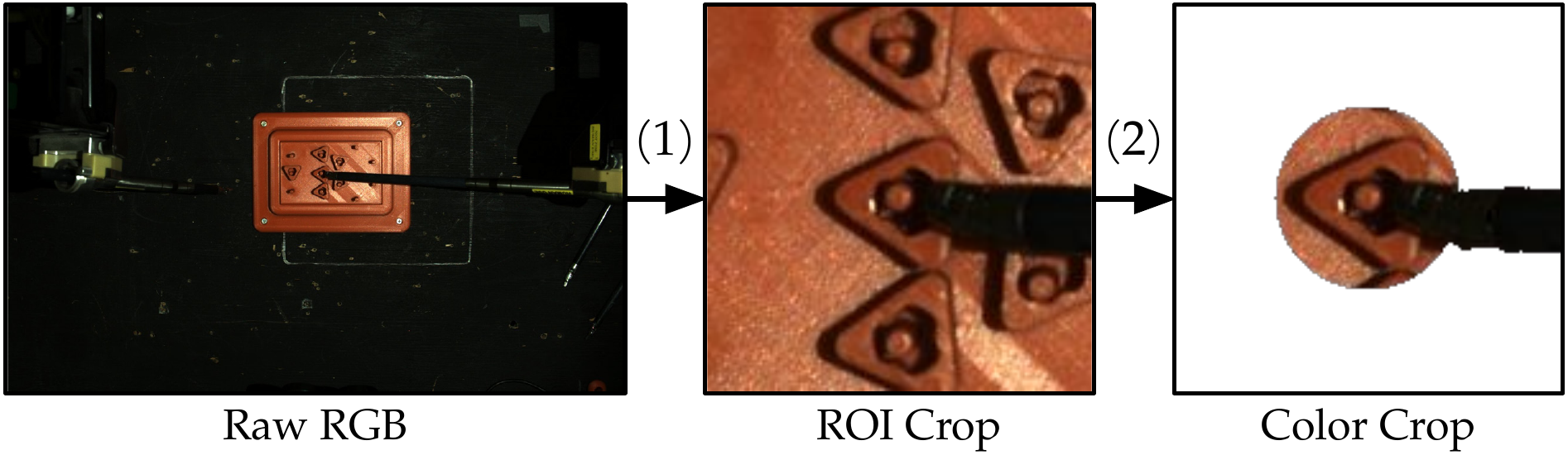}
\caption{
\textbf{Data Filtering}. We preprocess the images in two ways: (1) crop a 150$\times$150 image around the center of the target peg, and (2) color-crop out all red pixels outside of a block-sized radius from the center of the peg.
} 
\label{fig:data-filtering}
\end{figure}

To train a single model on all 180 demonstrations, regardless of the target peg, we (1) crop images around the peg, and (2) color crop other blocks (see Fig.~\ref{fig:data-filtering}).
For (1), we crop a 150$\times$150 image centered on the target peg.
For (2), we color-crop out all red pixels outside of a block-sized radius from the center of the target peg. This removes the other blocks from the input image as much as possible while keeping the instrument, target block, and target peg visible.
$I_t = \mathrm{Preprocess}( I^{\rm raw}_t)$ refers to the image $I^{\rm raw}_t$ after preprocessing (see Fig.~\ref{fig:data-filtering}).

\begin{figure}[t]
\centering
\includegraphics[width=1.0\linewidth]{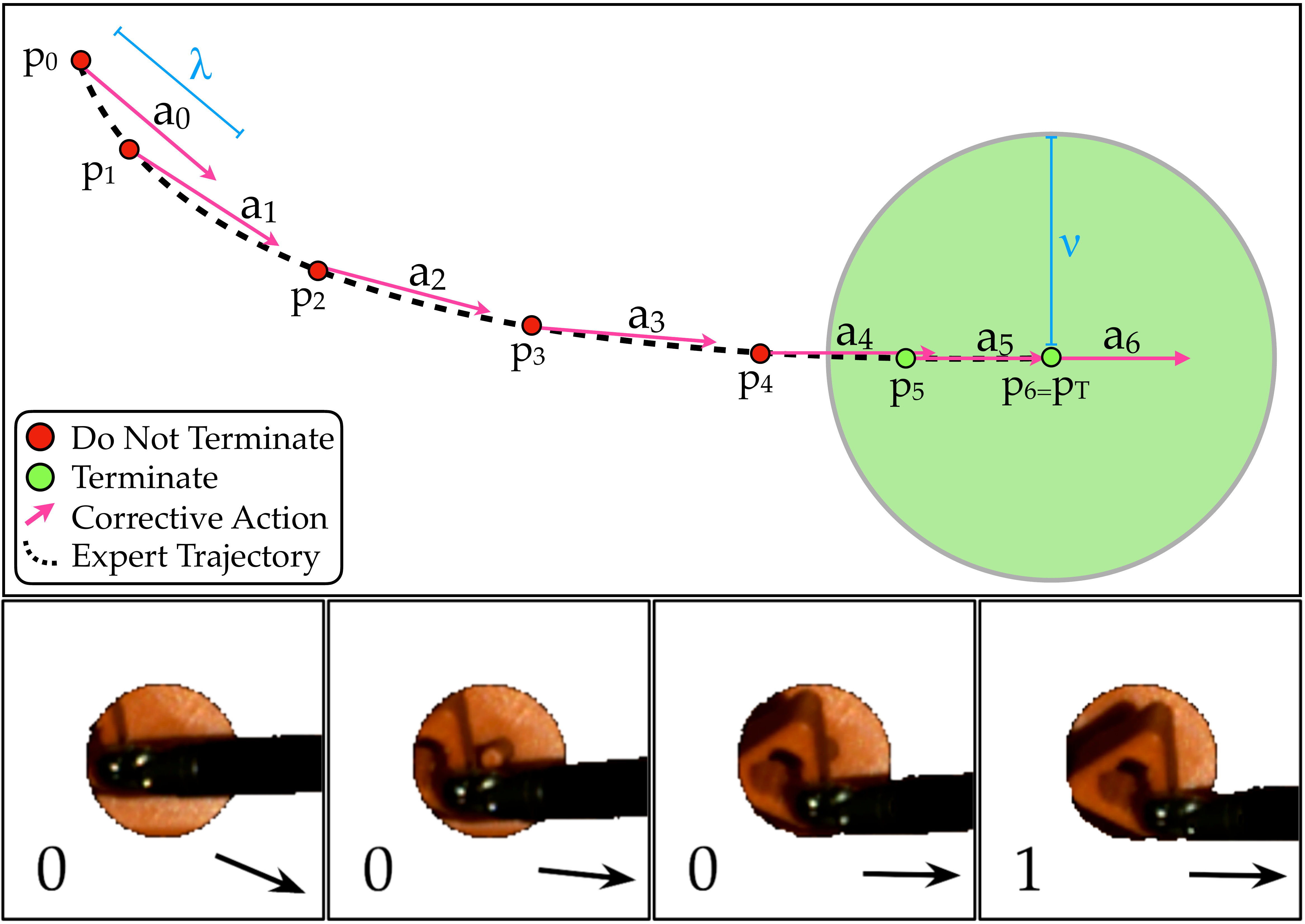}
\caption{
\textbf{Supervision Extraction with Labeled Example.}  \textbf{Top:} The action $a_t$, indicated by the arrows, is a vector of length 1~mm~($\lambda$) in the direction $p_{t'} - p_t$ traveled by the expert trajectory, where $p_t$ is the current waypoint and $p_{t'}$ is the next waypoint that is at least 1~mm~($\lambda$) distance from $p_t$. The extracted termination label, indicated by the color of the waypoint, is $1$ if the distance to the final position in the trajectory $p_T$ is less than 2~mm~($\nu$), and $0$ otherwise. Each waypoint corresponds to an image, and each image receives both a corrective action label and a termination signal label. \textbf{Bottom:} Preprocessed images (see Fig.~\ref{fig:data-filtering}) from a corrective place trajectory with labels extracted using method described above. 
}
\label{fig:training-traj-example}
\end{figure}

\begin{figure*}[t]
\centering
\includegraphics[width=1.0\linewidth]{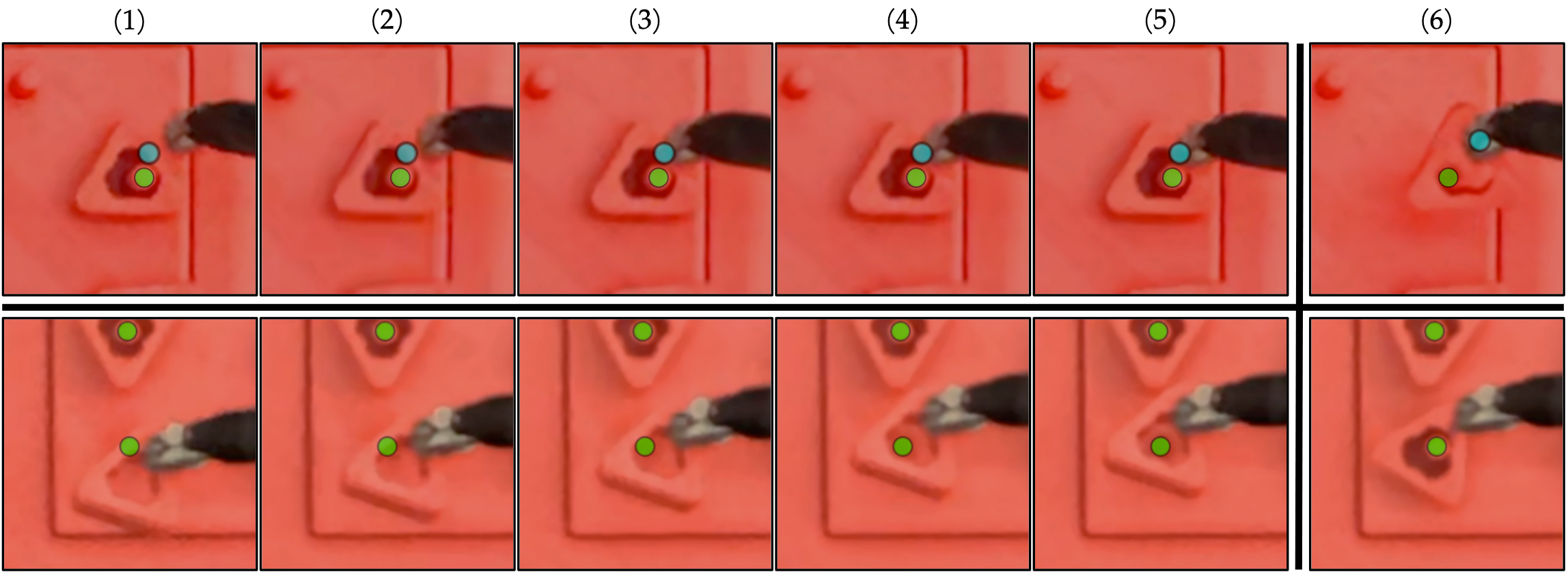}
\caption{
\textbf{IVS Example.} Example of IVS correcting positioning errors. \textbf{Top row (pick subtask)}: IVS removes pickup error with 5 corrective updates in 0.6 seconds. We overlay small blue circles to highlight an optimal pick location on the block. The positioning of the robot is off due to the inaccuracy of the coarse policy, as the end-effector is not positioned over the block (Frame 1). We then switch to the learned policy, and visual servoing guides the end-effector over a pick point (Frames 2-5), and once determined a safe pickup is possible, picks the block successfully (Frame 6). \textbf{Bottom~row~(place~subtask)}:  IVS removes placement error with 12 corrective updates in 1.2 seconds. We overlay small green circles to highlight the location of the pegs. The positioning of the robot is off due to the inaccuracy of the coarse policy, as the peg is not under the block (Frame 1). We then switch to the learned policy, and visual servoing guides the block over the peg (Frames 2-5), and once determined a safe situation to drop, places the block successfully (Frame 6).
} 
\label{fig:pick-n-place-example}
\end{figure*}

\subsubsection{Supervision Extraction}\label{subsec:labeling}

We additionally process the datasets to extract supervision for subsequent learning. For an input raw trajectory $\mathcal{T}^{\rm raw}$, we transform it via mappings $\Pi_{\rm action}$ and $\Pi_{\rm term}$ that extract the action executed and a terminal condition, respectively (see Fig.~\ref{fig:training-traj-example}).

\textbf{Corrective Action Extraction:}
We transform $\mathcal{T}^{\rm raw}$ to get the action-labeled dataset $\mathcal{T_{\rm action}} = \Pi_{\rm action}(\mathcal{T}^{\rm raw}) = \left\{(I_t, a_t)\right\}_{t=1}^{T}$, where
\begin{align*}
    a_t &= \lambda\frac{p_{t'} - p_t}{\|(p_{t'} - p_t)\|_2}\\
    \text{ s.t. } t' &= \min\left( \left\{t'' \mid t''>t \wedge \lVert p_{t''} - p_t\rVert_2 \geq \lambda\right\} \cup \{T\}\right).
\end{align*}
The action $a_t$ is a vector of length $\lambda$ in the direction $p_{t'} - p_t$ traveled by the demonstration trajectory, where $p_{t'}$ is the next waypoint that is at least $\lambda$ distance from $p_t$. In the case the distance to the final position in the trajectory $p_T$ is less than $\lambda$, $p_{t'} = p_T$.
In experiments, we set the $\lambda$ hyperparameter to 1~mm, as 1~mm is an upper bound on the expected Cartesian distance traveled by the end-effector per corrective update. 

\textbf{Terminal Condition Extraction:}
We extract a binary completion label for each image. We transform $\mathcal{T}^{\rm raw}$ to get the action-labeled dataset $\mathcal{T_{\rm term}} = \Pi_{\rm term}(\mathcal{T}^{\rm raw}) = \left\{(I_t, \phi_t)\right\}_{t=1}^{T}$, where
$\phi_t = \mathbbm{1}\left\{\|p_{T} - p_t\|_2 \leq \nu\right\}$.
The flag $\phi_t$ is $1$ if the distance to the final position $p_T$ is less than the hyperparameter $\nu$.  In experiments, we set $\nu$ to 2~mm, as 2~mm is strict enough to reliably confirm termination, and lenient enough to prevent label imbalance, as 30\% of images are labeled positively. 


\subsection{Constructing the Visual Feedback Policy}

We train the visual feedback policy $\pi_{VS}$ for each subtask from demonstrations using supervised learning. The policy takes in a top-down RGB image $I_t$ as input and outputs $(a_t, \phi_t)$, where $a_t$ is an action vector and $\phi_t$ is a termination condition.

\subsubsection{Training the Visual Feedback Policy}
The policy consists of an ensemble of 4 Convolutional Neural Networks~\cite{alexnet,Lecun98gradient-basedlearning}, denoted by 
$f_{\theta_{[1 .. 4]}}$. 
Each individual model $f_{\theta_i}$ consists of alternating convolution and max pooling layers, following by dense layers separated by Dropout~\cite{dropout}. 
We use an ensemble of $k$ models to make the policy more robust, and we evaluate for $k \in \{1,2,4,8\}$ in Table~\ref{tab:ensembles}. We select $k$ to be $4$. Each model $f_{\theta_i}$ uses a processed $150\times150\times3$ image $I_t$ as input and outputs estimates $a_{t,i}$ and $\phi_{t,i}$ of the supervisor action and terminal conditions respectively. Each model trains on 150 randomly sampled trajectories, with 30 for testing. We train each network by minimizing $l_{\rm MSE} + \mu l_{\rm CE}$ on sampled batches of its training data, where $l_{\rm MSE}$ is a Mean Square Error loss on the action prediction, $l_{\rm CE}$ is a cross-entropy loss for the terminal condition prediction, and $\mu$ is a relative weighting hyperparameter. The weights in the convolutional layers are shared, as useful convolutional filters are likely similar across subtasks, while the weights in the dense layers are independent. Sharing convolutional layers provides two sources of supervision when training the filters.

\begin{table}[t]
\caption{
\small
\textbf{Ablation Study Varying Number of Models in Ensemble}. We compare IVS with differing number of models in the ensemble across $3$ full trials of peg transfer (36 transfers). More models results in a more robust policy, but due to compute limitations, less frequent servoing. More frequent servoing results in higher precision, as both the corrective action and the termination signal are updated more frequently. Thus, the goal is to find a balance between maximizing robustness (many models) and maximizing update frequency (few models). In this work, we use an ensemble of 4 models. 
}
\centering
\begin{tabular}{| l || c | c | c |}
\hline \small
{\footnotesize Num Models} & Update Frequency & Transfer Success Rate 
\\ \hline\hline 
1 & 15.6 & 97.2\% 
\\ \hline
2 & 12.8 & 98.6\% 
\\ \hline
\textbf{4} & \textbf{10.0} & \textbf{100.0\%} 
\\ \hline
8 & 7.1 & 100.0\% 
\\ \hline
\end{tabular}
\vspace{2pt}
\label{tab:ensembles}
\end{table}

\begin{table*}[t]
\caption{
\small
\textbf{Peg Transfer Baseline Comparison}. Benchmark comparing performance of IVS to the baselines described in Section~\ref{sec:exps}. IVS beats both baselines in terms of pick success rate, place success rate, and overall transfer success rate. Due to using RGB imaging, we are able to take many corrective steps per second without stopping, minimizing additions to the mean transfer time.
}

\centering
\begin{tabular}{| l || r | r || r || r | r |}
\hline 
 & Pick Success Rate & Place Success Rate & Mean Transfer Time (s) & Success / Attempts & Transfer Success Rate \\ \hline\hline 
Uncalibrated Baseline & 96.3\% & 75.7\% & \textbf{8.7} & 77 / 106 & 72.6\% \\ \hline
Calibrated Baseline &  97.9\% & 99.6\% & 9.5 & 116 / 119 & 97.5\% \\ \hline
IVS & \textbf{99.2\%} & \textbf{100\%} & 10.2 & 118 / 119 & \textbf{99.2\%} \\ \hline

\end{tabular}
\vspace{10pt}
\label{tab:accuracy}
\end{table*}

\subsubsection{Querying the Visual Feedback Policy}
Once trained, we evaluate the ensemble of models in parallel with a filtered RGB image (Sec.~\ref{subsubsec:image-filtering}). We let
\begin{align*}
    \pi_{VS}(I_t) &= (a_t, \phi_t) = \left(\frac{1}{4} \sum_{i=1}^{4}a_{t,i},\; \sum_{i=1}^{4}\mathbbm{1}\left\{\phi_{t,i} \geq \omega \right\} \geq \kappa\right),
\end{align*}



%
where $\omega$ and $\kappa$ are a hyperparameters set to 0.70 and 3 respectively. We hand-tune these hyperparameters to maximize speed and minimize false positives. The predicted action is the mean action across the ensemble, and the predicted termination condition checks if at least $\kappa$ models predict termination with probability greater than $\omega$.

\section{Experiments}\label{sec:exps}

The experimental setup has a top-down RGBD camera and uses the teleoperation interface~\cite{dvrk2014} to collect demonstrations on the dVRK. Training data are collected on a single instrument, but the system is tested with $3$ different instruments that have unique dynamics due to differences in cabling properties. We use the position of the blocks and pegs estimated by an RGBD image to construct trajectories for picks and places, but only use RGB images for visual servoing.

We benchmark IVS against two baselines:
\begin{itemize}
    \item \textbf{Uncalibrated Baseline (UNCAL)}: This is a coarse open-loop policy, implemented using the default unmodified dVRK controller. The trajectories are tracked in closed-loop with respect to the robot's odometry, but open-loop with respect to vision. 
    
    \item \textbf{Calibrated Baseline (CAL)}: This is a calibrated open-loop policy~\cite{hwang2020efficiently} that is the current state-of-the-art method for automating peg transfer.
    To correct for backlash, hysteresis, and cable tension, the authors train a recurrent dynamics model to estimate the true position of the robot based on prior commands. Similar to the uncalibrated baseline, the robot tracks reference trajectories in closed-loop with respect to the position estimated by the recurrent model, but open-loop with respect to visual inputs.
\end{itemize}
\begin{table}[t]
\centering
\caption{
\small
\textbf{IVS Efficiency Benchmark.} We analyze efficiency of IVS on 3 large needle driver instruments (see Fig.~\ref{fig:main}). IVS adds an additional 1.22 seconds per transfer. 
}
\begin{tabular}{| l || c | c | c | }
\hline
\multicolumn{4}{|c|}{Pick}\\ \hline
Instrument & Corrective Updates & Time (sec) & Distance (mm)\\ \hline\hline
A & 5.75 $\pm$ 2.10 & 0.60 $\pm$ 0.19 & 1.71 $\pm$ 0.65 \\\hline
B & 7.73 $\pm$ 2.63 & 0.77 $\pm$ 0.22 & 2.52 $\pm$ 0.78 \\\hline
C & 7.43 $\pm$ 3.03 & 0.74 $\pm$ 0.26 & 2.24 $\pm$ 0.85 \\\hline
Mean & 7.01 $\pm$ 2.77 & 0.70 $\pm$ 0.24 & 2.17 $\pm$ 0.84 \\
\hline
\multicolumn{4}{c}{\vspace{-3pt}}\\
\hline
\multicolumn{4}{|c|}{Place}\\ \hline
Instrument & Corrective Updates & Time (sec) & Distance (mm)\\
\hline\hline
A & 4.90 $\pm$ 4.82 & 0.52 $\pm$ 0.43 & 2.04 $\pm$ 1.77 \\ \hline
B & 5.63 $\pm$ 4.20 & 0.58 $\pm$ 0.37 & 2.12 $\pm$ 1.57 \\ \hline
C & 4.68 $\pm$ 4.21 & 0.49 $\pm$ 0.37 & 1.78 $\pm$ 1.43 \\\hline
Mean & 4.97 $\pm$ 4.47 & 0.52 $\pm$ 0.39 & 1.93 $\pm$ 1.58 \\\hline
\end{tabular}
\vspace{10pt}
\label{tab:timing}
\end{table}

\subsection{Accuracy Results}

Overall, IVS achieves high success rates in the peg transfer task, with a pick and place success rate of 99.2\% and 100.0\% respectively. IVS succeeds on 118 of 119 transfers, resulting in a 99.2\% transfer success rate, exceeding the uncalibrated baseline by over 25\%. See Table~\ref{tab:accuracy} for details and we illustrate an example of IVS correcting positioning errors in Figure~\ref{fig:pick-n-place-example}. 

\subsection{Timing Results}

The goal is to produce higher success rates, rather than to reduce the timing. However, we find that the proposed method is only marginally slower than the baselines (see Table~\ref{tab:accuracy}). Due to fast image capture and continuous servoing via RGB imaging, we are able to both update the robot's velocity and check for termination 10 times per second, minimizing additions to the mean transfer time. As a result, the mean transfer time is only 1.5~seconds slower than the uncalibrated baseline, and 0.7~seconds slower than the calibrated baseline.

We report IVS timing results in Table~\ref{tab:timing}. On average, each pick requires 2.2~mm of correction, spanning 0.7~seconds and 7 corrective updates, and each place requires 1.9~mm of correction, spanning 0.5~seconds and 5 corrective updates. 

\subsection{Transferability Results}\label{subsec:transferability}

To conduct instrument transfer evaluation, we experiment using multiple large needle drivers. Each instrument has inconsistent cabling characteristics resulting in various success rates.
We have 3 different instruments: \emph{A}, \emph{B}, and \emph{C} (Fig.~\ref{fig:main} bottom).  We trained with data from instrument \emph{A}.

We investigate whether models learned from data using one instrument can transfer to another instrument without modification. This is challenging, because different surgical instruments, even of the same type, have different cabling properties due to differences in wear and tear. However, the visual servoing algorithm does not rely on the cabling characteristics of any specific instrument, but rather only requires that the robot is able to roughly correct in the desired direction. We hypothesize that errors in executing the corrective motion can be mitigated over time by executing additional corrective motions, as long as the cumulative error is decreasing. However, the calibrated baseline uses an observer model that explicitly predicts the motion of the robot based on prior commands, which requires learning the dynamics of the specific instrument used in training which may not be sufficiently accurate on a new instrument. We report transferability results in Table~\ref{tab:transfer} and observe that the IVS model trained on Instrument \emph{A}
does not decrease in performance on different instruments, while the calibrated baselines suffer significantly on different instruments.

\begin{table}[t]
\caption{
\small
\textbf{Instrument Transfer Comparison}. Benchmark comparing performance on 10 full trials of peg transfer (120 transfers) of the uncalibrated baseline, calibrated baseline, and IVS across 3 different surgical instruments with unique cabling characteristics. IVS consistently beats both baselines, while remaining robust to instrument changes.
}
\centering
\begin{tabular}{| l || c || c | c | c || c |}
\hline 
Instrument & UNCAL & CAL$_\text{A}$ & CAL$_\text{B}$ & CAL$_\text{C}$ & IVS$_\text{A}$ \\ 
\hline\hline 
A & 72.6\% 
& 97.5\% 
& 48.1\% & 55.2\% & \textbf{99.2\%} 
\\\hline 
B & 31.3\% 
& 58.5\% & 98.3\% & 67.0\% & \textbf{99.2\%} 
\\\hline 
C & 47.2\%
& 27.8\% & 81.6\% & 97.4\% &
\textbf{100.0\%} 
\\\hline\hline 
Mean & 50.5\% & 69.8\% & 77.3\% & 79.2\% &
\textbf{99.4\%} 
\\\hline 
\end{tabular}
\vspace{10pt}
\label{tab:transfer}
\end{table}





\section{Discussion and Future Work}

In this work, we propose intermittent visual servoing to attain the highest published success rates for automated surgical peg transfer. IVS maintains performance across different instruments surprisingly well (see Table~\ref{tab:transfer}), and this transferability is critical to implementing automated surgical techniques where instruments are being exchanged frequently and a instrument's properties are expected to change over time.
In future work, we will investigate how to further decrease timing, perform automated bilateral peg transfer with handovers between $2$ arms in each transfer, and apply IVS to surgical cutting~\cite{thananjeyan2017multilateral}, surgical suturing~\cite{sen2016automating}, and non-surgical applications such as assembly~\cite{lozano1976design}.

\section*{Acknowledgements}

This research was performed at the AUTOLAB at UC Berkeley in affiliation with the Berkeley AI Research (BAIR) Lab, Berkeley Deep Drive (BDD), the Real-Time Intelligent Secure Execution (RISE) Lab, the CITRIS ``People and Robots'' (CPAR) Initiative, and with UC Berkeley's Center for Automation and Learning for Medical Robotics (Cal-MR). This work is supported in part by the Technology \& Advanced Telemedicine Research Center (TATRC) project W81XWH-18-C-0096 under a medical Telerobotic Operative Network (TRON) project led by SRI International. The authors were supported in part by donations from Intuitive Surgical, Siemens, Google, Toyota Research Institute, Honda, and Intel. The da Vinci Research Kit was supported by the National Science Foundation, via the National Robotics Initiative (NRI), as part of the collaborative research project ``Software Framework for Research in Semi-Autonomous Teleoperation'' between The Johns Hopkins University (IIS 1637789), Worcester Polytechnic Institute (IIS 1637759), and the University of Washington (IIS 1637444).  Daniel Seita is supported by a Graduate Fellowships for STEM Diversity.


\bibliographystyle{IEEEtranS}
\bibliography{IEEEabrv, reference}


\end{document}